\documentclass{article}

\usepackage{PRIMEarxiv}
\usepackage{times}  % DO NOT CHANGE THIS
\usepackage{amssymb}
\usepackage{amsmath}
\usepackage{booktabs}
\usepackage{multirow}
\usepackage[utf8]{inputenc} % allow utf-8 input
\usepackage[T1]{fontenc}    % use 8-bit T1 fonts
\usepackage{hyperref}       % hyperlinks
\usepackage{url}            % simple URL typesetting
\usepackage{booktabs}       % professional-quality tables
\usepackage{amsfonts}       % blackboard math symbols
\usepackage{nicefrac}       % compact symbols for 1/2, etc.
\usepackage{microtype}      % microtypography
\usepackage{lipsum}
\usepackage{fancyhdr}       % header
\usepackage{graphicx}       % graphics
\graphicspath{{media/}}     % organize your images and other figures under media/ folder

\title{Spatial-Functional awareness Transformer-based graph archetype contrastive learning for Decoding Visual Neural Representations from EEG
\thanks{\textit{\underline{Citation}}: 
\textbf{Authors. Title. Pages.... DOI:000000/11111.}} 
}

\author{
  Yueming Sun \\
  Hybrid Intelligence Lab \\
  Durham University \\
  Durham\\
  \texttt{yueming.sun@durham.ac.uk} \\
   \And
  Long Yang \\
  Hybrid Intelligence Lab \\
  Durham University \\
  Durham\\
  \texttt{long.yang@durham.ac.uk} \\
}

\begin{document}
\maketitle

\begin{abstract}
Decoding visual neural representations from Electroencephalography (EEG) signals remains a formidable challenge due to their high-dimensional, noisy, and non-Euclidean nature. In this work, we propose a Spatial-Functional Awareness Transformer-based Graph Archetype Contrastive Learning (SFTG) framework to enhance EEG-based visual decoding. Specifically, we introduce the EEG Graph Transformer (EGT), a novel graph-based neural architecture that simultaneously encodes spatial brain connectivity  and temporal neural dynamics. To mitigate high intra-subject variability, we propose Graph Archetype Contrastive Learning (GAC), which learns subject-specific EEG graph archetypes to improve feature consistency and class separability. Furthermore, we conduct comprehensive subject-dependent and subject-independent evaluations on the Things-EEG dataset, demonstrating that our approach significantly outperforms prior state-of-the-art EEG decoding methods.The results underscore the transformative potential of integrating graph-based learning with contrastive objectives to enhance EEG-based brain decoding, paving the way for more generalizable and robust neural representations.
\end{abstract}

\section{Introduction}
Recent advances in deep learning have revolutionized both computer vision and brain–computer interface (BCI) research. In computer vision, transformer‐based models and large-scale pre-training methods—exemplified by Vision Transformer \cite{dosovitskiy2020image}, CLIP \cite{radford2021learning}, and OmniVL \cite{wang2022omnivl}—have set new performance standards across a wide range of tasks. In parallel, EEG decoding has rapidly advanced through architectures that learn directly from raw brain signals \cite{lawhern2018eegnet, song2022eeg, ding2023lggnet}. However, these two domains have evolved largely independently: while computer vision models excel at visual recognition, BCI systems decode neural activity with high temporal fidelity yet often fail to leverage the rich spatial–semantic information inherent in images.

Electroencephalography offers an unparalleled window into the rapid neural dynamics underlying human cognition. Recent EEG-based research has demonstrated its potential not only for understanding visual processing but also for enhancing brain–computer interfaces. For instance, VigilanceNet \cite{cheng2022vigilancenet} decouples intra- and inter-modality learning to robustly estimate vigilance levels in RSVP-based paradigms, while graph attention networks \cite{petar2018graph} provide powerful mechanisms for capturing complex spatial dependencies in EEG data. Studies on the temporal hierarchy of object processing in human visual cortex \cite{xu2023temporal} further highlight EEG’s unique ability to resolve neural representations on a millisecond scale. Moreover, aligning artificial model representations with biological data from primate visual cortex has shown promise in improving behavioral predictability and adversarial robustness \cite{dapello2022aligning}. The recent release of a large and richly annotated EEG dataset for modeling human visual object recognition \cite{gifford2022large} has also opened new avenues for data-driven approaches that bridge the gap between artificial neural networks and biological vision.

EEG decoding has evolved through the integration of advanced deep learning techniques and rigorous experimental design. Recent approaches, such as MAtt \cite{pan2022matt}, leverage manifold attention mechanisms to exploit the Riemannian geometry of EEG covariance matrices, thereby enabling robust extraction of spatiotemporal features. Complementary methods, like those proposed by Kobler et al. \cite{kobler2022spd}, introduce SPD domain-specific batch normalization to achieve interpretable unsupervised domain adaptation across sessions and subjects, addressing the notorious variability in EEG recordings. Meanwhile, multimodal learning frameworks that jointly model neural activity and visual features have yielded richer brain representations that enhance decoding performance \cite{palazzo2020decoding}. Nevertheless, the reliability of EEG classification is challenged by experimental pitfalls; Li et al. \cite{li2020perils} critically highlight how traditional block design experiments can confound results, while Ahmed et al. \cite{ahmed2021object} show that object classification from randomized EEG trials often achieves only marginal accuracy above chance when such confounds are eliminated. Together, these studies underscore the need for robust, geometrically informed, and carefully controlled methodologies in EEG decoding to advance BCI applications.

Finally, recent advances in neuroimaging and deep learning have enabled the direct decoding of visual information from human brain signals. Deep learning models can now classify visual categories from EEG with high temporal precision \cite{spampinato2017deep}, while large-scale multimodal datasets like THINGS-data \cite{hebart2023things} provide rich resources for exploring object representations across imaging modalities. High-temporal-resolution studies reveal that neural processing of objects unfolds within the first few hundred milliseconds \cite{cichy2014resolving}, and innovative approaches are reconstructing complex images from brain activity \cite{lin2022mind} and decoding both seen and imagined objects using hierarchical visual features \cite{horikawa2017generic}. Together, these developments underscore the transformative potential of integrating deep learning with neuroimaging data to unravel the mechanisms of human visual processing.

\noindent Our key contributions can be summarized as follows:
\begin{itemize}
\item We propose an EEG Graph Transformer (EGT) that integrates spatial brain connectivity (adjacent EEG electrodes) and temporal neural dynamics (functional interactions across brain regions) into a unified graph-based representation.
\item To address the high intra-subject variability in EEG signals, we introduce Graph Archetype Contrastive Learning (GAC), which constructs EEG graph archetypes for each subject based on their learned feature representations.
\item We conduct subject-dependent and subject-independent evaluations on the Things-EEG dataset. Our method outperforms prior approaches and demonstrating the effectiveness of our model in real-world EEG decoding tasks.
\end{itemize}

\section{Related Works}
\begin{figure}
    \centering
    \includegraphics[width=0.95\linewidth]{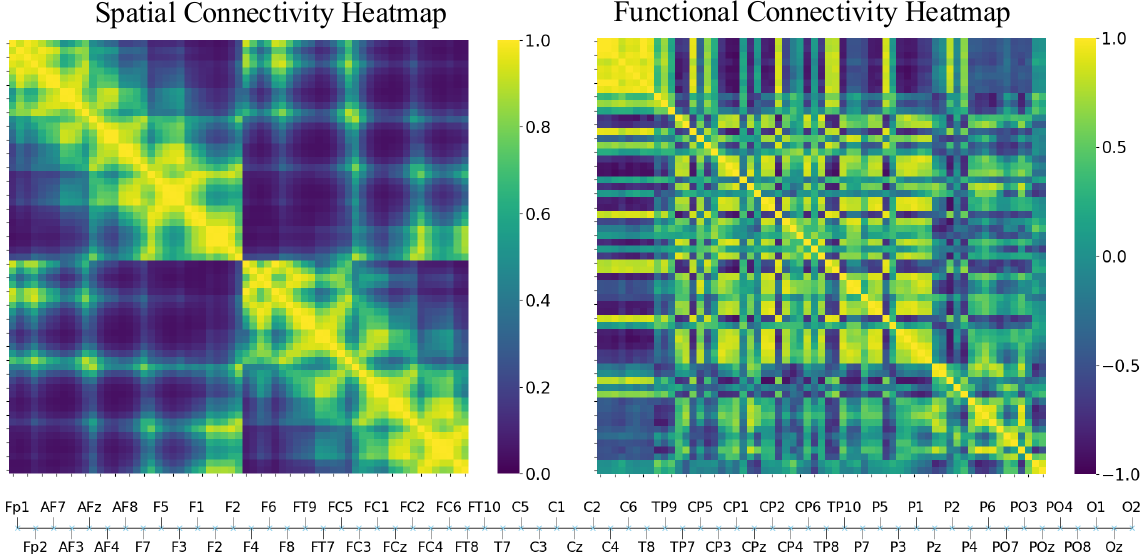}
    \caption{Spatial and functional connectivity across 63 EEG channels is quantified by calculating the Pearson correlation coefficient.}
    \label{fig:enter-label}
\end{figure}

\subsection{Brain decoding}
Recent advances in brain decoding have leveraged deep learning and neuroimaging techniques to reconstruct visual experiences from human brain activity. Traditional methods employed linear models and Bayesian approaches to infer visual stimuli from fMRI signals~\cite{miyawaki2008visual}. More recently, deep neural networks (DNNs) have been utilized for hierarchical feature extraction, enabling more accurate reconstructions~\cite{shen2019deep}. Studies have demonstrated the effectiveness of deep image reconstruction, where DNN feature decoding is combined with optimization algorithms to iteratively refine images to match decoded brain representations~\cite{takagi2023high}. Moreover, latent diffusion models have emerged as a promising approach, significantly improving the quality of reconstructions by capturing fine-grained details in neural responses~\cite{takagi2023high}. Despite these advancements, challenges remain in generalizing across subjects due to individual variability in brain organization. Functional alignment techniques, such as neural code conversion, aim to mitigate these differences by translating neural responses between individuals~\cite{ho2023inter}. These developments pave the way for broader applications of brain decoding, including brain-computer interfaces (BCI) and cognitive state monitoring. In the context of BCI, advances in SSVEP-based systems have improved performance through task-discriminant component analysis~\cite{liu2021improving} and representative-based cold-start methods~\cite{shi2023representative}. Additionally, high-performance brain-to-text communication has been demonstrated using handwriting-based decoding methods~\cite{willett2021high}. These findings highlight the growing impact of deep learning and advanced neural decoding strategies in bridging the gap between human cognition and machine interpretation.

\begin{figure*}[t]
    \centering
    \includegraphics[width=0.95\linewidth]{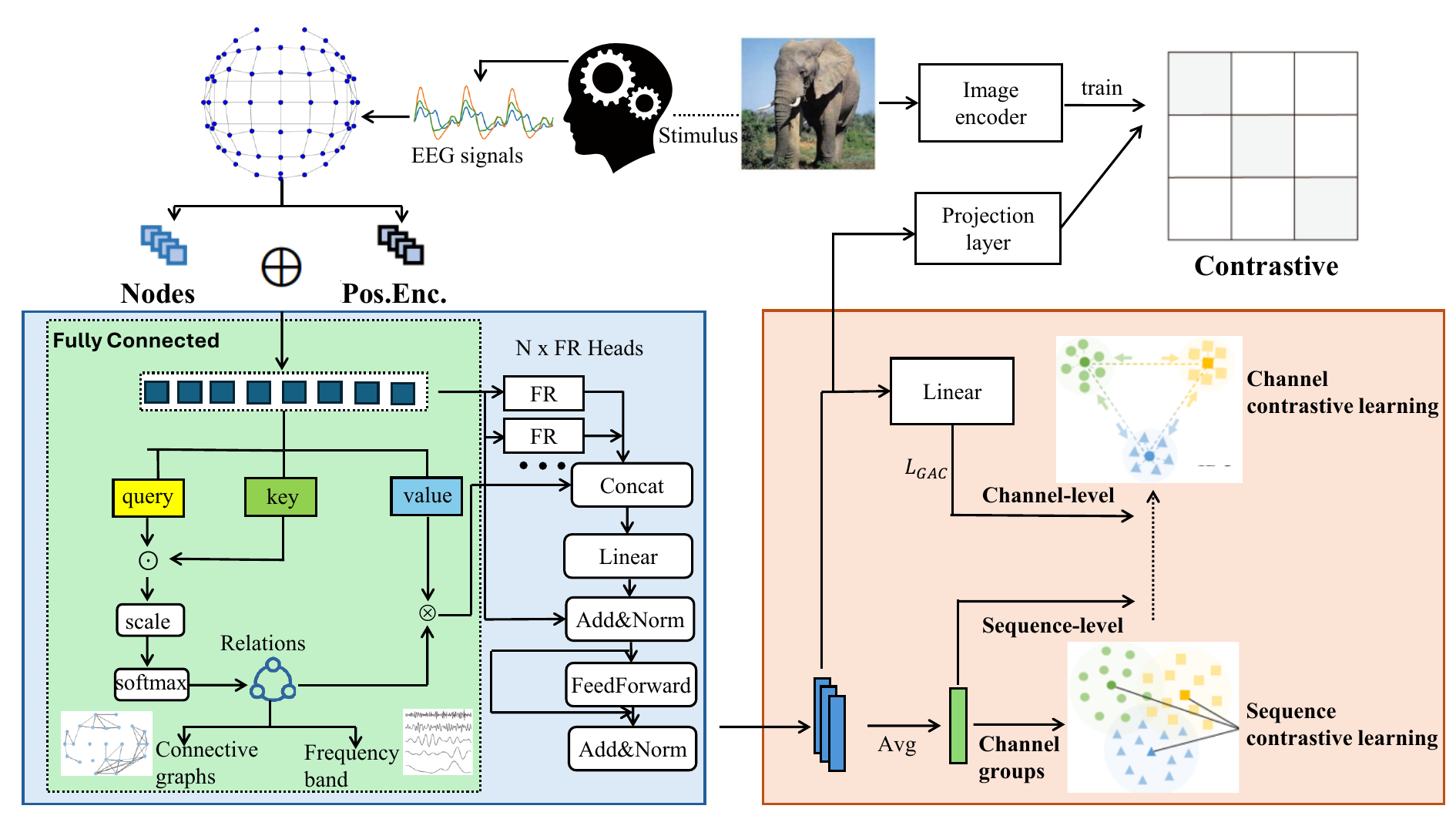}
    \caption{Our framework consists of the EEG Graph Transformer (EGT) (blue box) and Graph Archetype Contrastive Learning (GAC) (orange box). We construct EEG graphs by leveraging spatial and functional connectivity, transforming raw EEG signals into node and graph embeddings with additional positional encodings. the outputs of the FR heads are integrated into transformer encoder to enhance feature extraction. The encoded EEG features are projected and trained to align with visual representations.}
    \label{fig:enter-label}
\end{figure*}

\subsection{Graph neural networks}
The application of graph neural networks (GNNs) has evolved through multiple methodological innovations in EEG-based emotion recognition. Pioneering work by \cite{song2018eeg} introduced dynamical graph convolutional neural networks (DGCNN), which explicitly modeled time-varying functional connectivity between EEG channels through adaptive adjacency matrices. To address overfitting in small EEG datasets, \cite{zhong2020eeg} proposed regularized graph neural networks (RGNN) incorporating neurophysiological priors such as inter-hemispheric asymmetry and brain network sparsity. Subsequent research explored hybrid architectures: \cite{song2021graph} developed a graph-embedded CNN framework that jointly learned spatial topologies from EEG scalp maps and spectral features via multi-branch fusion. Further advancing personalization, \cite{song2021variational} devised variational instance-adaptive graphs (V-IAG) that generated subject-specific neural representations through amortized inference. Recent work by \cite{demir2021eeg} established EEG-GNN, an end-to-end architecture optimizing both graph construction (based on spectral coherence) and temporal feature extraction using dilated convolutions. However, despite these advances, graph-based approaches remain underexplored in EEG-to-image decoding tasks, where reconstructing visual stimuli from brain activity requires modeling hierarchical spatiotemporal interactions across distributed cortical networks.

\section{Methodology}
\subsection{Problem formulation}

Electroencephalography (EEG)  signals can be represented as a matrix \(X \in \mathbb{R}^{C \times T}\), where \(C\) denotes the number of channels, and \(T\) represents the number of time steps. Unlike image data, where pixels are arranged in a structured 2D grid, EEG electrodes are distributed in a discrete, non-Euclidean manner across the scalp. The spatial relationships among EEG channels are better characterized by a graph, defined as:

\begin{equation}
\mathbf{A} = \{ A_{ij} \} \in \mathbb{R}^{C \times C}
\end{equation}

where \(\mathbf{A}_{ij}\) encodes the connectivity strength between channels \(i\) and \(j\). This adjacency matrix can be defined based on physical distance, functional connectivity, or statistical dependencies. We establish the following two key assumptions: \textbf{Spatial Discontinuity}: EEG electrodes are distributed non-uniformly, making it difficult to define consistent local neighborhoods in a Euclidean space. \textbf{Functional Connectivity}: Non-adjacent EEG channels may exhibit strong correlations\cite{stam2014modern} (e.g.,inter-hemispheric synchronization), requiring a global interaction modeling approach.

Convolutional Neural Networks (CNNs) rely on localized, structured spatial dependencies, which may not align with EEG topology. The standard 2D convolution operation is defined as:
\begin{equation}
Y_{c'} = \sum_{k=1}^{K} W_{c',k} \ast X_k + b_{c'}
\end{equation}

where \(W_{c',k}\) represents a fixed-size convolutional kernel (e.g., \(3 \times 3\)). This operation assumes a structured spatial layout, which is not applicable to EEG channels. When EEG electrodes are projected onto a 2D plane for convolutional processing, physically distant but functionally related electrodes may fall within the same receptive field, leading to feature extraction inconsistencies. Moreover, increasing the receptive field in CNNs requires stacking multiple layers, which results in an exponential growth in the number of parameters and computational cost \cite{kipf2017semi}. This limitation makes CNNs suboptimal for learning complex, long-range dependencies in EEG data.

Graph Neural Networks (GNNs) overcome these limitations by leveraging an explicit graph structure that models arbitrary relationships among EEG channels. The Graph Convolutional Network (GCN) is formulated as:
\begin{equation}
 \mathbf{H}^{(l+1)} = \sigma\left( \tilde{\mathbf{D}}^{-\frac{1}{2}} \tilde{\mathbf{A}}\, \tilde{\mathbf{D}}^{-\frac{1}{2}} \mathbf{H}^{(l)} \mathbf{W}^{(l)} \right)
\end{equation}

where \(\tilde{\mathbf{A}} = \mathbf{A} + \mathbf{I}\) (self-connections added), \(\tilde{\mathbf{D}}\) is the degree matrix of \(\tilde{\mathbf{A}}\), \(\mathbf{W}^{(l)}\) represents the learnable weight matrix, and \(\sigma(\cdot)\) is an activation function. This formulation enables each node (EEG channel) to aggregate information from its neighbors, capturing complex dependencies in a non-Euclidean space. Additionally, Graph Attention Networks (GATs) introduce an adaptive weighting mechanism, dynamically learning the importance of interactions between EEG channels:
\begin{equation}
 \alpha_{ij} = \text{softmax}_j\left( \text{LeakyReLU}\left( \mathbf{a}^T \left[ \mathbf{W}\mathbf{h}_i \, \Vert \, \mathbf{W}\mathbf{h}_j \right] \right) \right) 
\end{equation}

where \(\alpha_{ij}\) represents the attention coefficient between nodes \(i\) and \(j\), while \(\mathbf{W}\) and \(\mathbf{a}\) are learnable parameters. The concatenation operator \(\Vert\) ensures that node features are jointly learned during training. Unlike CNNs, GNNs naturally support non-local interactions and enable global feature aggregation in a single layer \cite{velivckovic2017graph}.
While CNNs tend to enforce spatial locality, leading to diagonalized covariance structures, GNNs directly model \(\mathbf{Cov}(\mathbf{X})\) via the adjacency matrix, preserving off-diagonal elements that encode inter-channel relationships. This property makes GNNs inherently better suited for tasks where cross-channel dependencies are crucial \cite{ortega2018graph}.

\subsection{EEG Graph Construction}
Based on several official EEG headset product documents \cite{easycap2018} \cite{biosemi64} \cite{openBCI}, we manually modeled brain activity recorded from multiple electrodes as a graph to capture intricate spatial and functional relationships among different brain regions.We construct EEG graphs by treating the electrode positions as nodes and defining edges based on functional or spatial connectivity. Each graph \( G_t(V_t, E_t) \), corresponding to the EEG signal at time step \( t \), consists of nodes: \( V_t = \{ v_t^1, v_t^2, \dots, v_t^J \}, v_t^i \in \mathbb{R}^d, i \in \{1, \dots, J\} \), and edges: \( E_t = \{ e_t^{i,j} \mid v_t^i, v_t^j \in V_t \}, e_t^{i,j} \in \mathbb{R} \).

Here, \( V_t \) and \( E_t \) represent the set of nodes corresponding to \( J \) EEG electrodes and the set of their connectivity relationships, respectively. The adjacency matrix of \( G_t \), denoted as \( A_t \in \mathbb{R}^{J \times J} \), encodes the pairwise relationships among the \( J \) nodes, where \( A_t(i, j) \) is set to 1 if electrodes \( i \) and \( j \) are connected, and 0 otherwise.\( A_t \) is initialized based on predefined spatial or functional connectivity patterns among the EEG electrodes.

\begin{table*}[t]
\centering
\caption{Overall accuracy of 200-way zero-shot classification: top-1 and top-5}
\label{tab:zeroshot}
\resizebox{\textwidth}{!}{
\begin{tabular}{l
                cc
                cc
                cc
                cc
                cc
                cc
                cc
                cc
                cc
                cc
                cc
                cc}
\toprule
\multirow{2}{*}{\textbf{Method}}
& \multicolumn{2}{c}{Subject1}
& \multicolumn{2}{c}{Subject2}
& \multicolumn{2}{c}{Subject3}
& \multicolumn{2}{c}{Subject4}
& \multicolumn{2}{c}{Subject5}
& \multicolumn{2}{c}{Subject6}
& \multicolumn{2}{c}{Subject7}
& \multicolumn{2}{c}{Subject8}
& \multicolumn{2}{c}{Subject9}
& \multicolumn{2}{c}{Subject10}
& \multicolumn{2}{c}{\multirow{2}{*}{Ave}}\\
\cmidrule(lr){2-21}
& \textbf{top-1} & \textbf{top-5}
& \textbf{top-1} & \textbf{top-5}
& \textbf{top-1} & \textbf{top-5}
& \textbf{top-1} & \textbf{top-5}
& \textbf{top-1} & \textbf{top-5}
& \textbf{top-1} & \textbf{top-5}
& \textbf{top-1} & \textbf{top-5}
& \textbf{top-1} & \textbf{top-5}
& \textbf{top-1} & \textbf{top-5}
& \textbf{top-1} & \textbf{top-5} \\
\midrule
\multicolumn{21}{c}{\textbf{Subject dependent --- train and test on one subject}} \\
BraVL\cite{du2023decoding}
& 6.1 & 17.9
& 4.9 & 14.9
& 5.6 & 15.1
& 4.0  & 13.4
& 6.0 & 18.2
& 6.5  & 20.4
& 8.8 & 23.7
& 4.3 & 14.0
& 7.0  & 19.7
& 7.0  & 19.7
& 5.8 & 17.5 \\

NICE\cite{song2024decoding}
& 12.3 & 36.6
& 13.1 & 39.0
& 16.4 & 47.0
& 8.0 & 26.9
& 14.1 & 40.6
& 15.2  & 42.1
& 20.0 & 49.9
& 13.3 & 37.1
& 14.9  & 41.9
& 14.9  & 41.9
& 13.8 & 39.5 \\

NICE-SA\cite{song2024decoding}
& 17.2 & 44.1
& 14.9 & 52.0
& 12.6 & 38.3
& 11.2 & 34.7
& 16.3 & 52.5
& 10.1 & 32.2
& 15.4 & 49.5
& 12.2 & 39.9
& 10.3 & 30.1
& 10.3 & 30.1
& 14.7 & 40.7 \\

NICE-GA\cite{song2024decoding}
& 18.5 & 45.0
& 15.5 & 52.7
& 13.2 & 38.8
& 11.7 & 35.6
& 17.1 & 53.3
& 10.5 & 33.0
& 16.0 & 50.2
& 12.8 & 40.3
& 11.1 & 30.8
& 11.1 & 30.8
& 15.6 & 41.2 \\

ATM-S~\cite{li2024visual} 
& 25.6 & 60.4 
& 22.0 & 54.5 
& 25.0 & 62.4 
& 31.4 & 60.9 
& 12.9 & 43.0 
& 21.3 & 51.1 
& 30.5 & 61.5 
& 38.8 & 72.0 
& 34.4 & 51.5 
& 29.1 & 63.5 
& 28.5 & 60.4 \\
VE-SDN~\cite{chen2024visual} 
& 32.6 & 63.7 
& 34.4 & 69.9 
& 38.7 & 73.5 
& 39.8 & 72.0 
& 29.4 & 58.6 
& 34.5 & 68.8 
& 34.5 & 68.3 
& 49.3 & 79.8 
& 39.0 & 69.6 
& 39.8 & 75.3 
& 37.2 & 69.9 \\
UBP \cite{Wu2025UBP}
& 41.2 & 70.5 
& 51.2 & 80.9 
& 51.2 & 82.0 
& 51.1 & 76.9 
& 42.2 & 72.8 
& 57.5 & 83.5 
& 49.0 & 79.9 
& 58.6 & 85.8 
& 45.1 & 76.2 
& 61.5 & 88.2 
& 50.9 & 79.7 \\

\textbf{SFTG(Ours)} 
& \textbf{45.6} & \textbf{75.7}
& \textbf{55.2} & \textbf{86.3}
& \textbf{54.1} & \textbf{85.2}
& \textbf{54.5} & \textbf{79.8}
& \textbf{46.6} & \textbf{76.9}
& \textbf{63.3} & \textbf{88.4}
& \textbf{56.1} & \textbf{84.7}
& \textbf{62.2} & \textbf{88.4}
& \textbf{48.5} & \textbf{82.3}
& \textbf{67.8} & \textbf{90.4}
& \textbf{55.4} & \textbf{83.8} \\

\midrule
\multicolumn{21}{c}{\textbf{Subject independent --- leave one subject out for test}} \\
BraVL \cite{du2023decoding}
& 7.2  & 20.1 
& 6.0  & 15.5 
& 5.9  & 14.7 
& 4.2  & 11.1 
& 7.8  & 22.6 
& 3.6  & 9.9 
& 6.2  & 17.8 
& 5.3  & 14.6 
& 4.4  & 13.0 
& 4.4  & 13.0 
& 5.6  & 15.5 \\
NICE \cite{song2024decoding}
& 8.1  & 22.8 
& 6.8  & 17.3 
& 6.4  & 15.9 
& 4.7  & 12.2 
& 8.6  & 23.5 
& 4.1  & 10.7 
& 7.0  & 19.3 
& 6.1  & 16.0 
& 4.9  & 14.2 
& 4.9  & 14.2 
& 6.2  & 21.4 \\
NICE-SA \cite{song2024decoding}
& 8.9  & 23.5 
& 7.3  & 17.9 
& 6.7  & 16.2 
& 4.9  & 12.6 
& 9.3  & 24.1 
& 4.4  & 11.4 
& 7.5  & 20.1 
& 6.5  & 16.7 
& 5.3  & 14.8 
& 5.3  & 14.8 
& 7.0  & 21.9 \\
NICE-GA \cite{song2024decoding}
& 8.2  & 23.0 
& 7.0  & 17.5 
& 6.4  & 15.7 
& 4.6  & 12.3 
& 8.5  & 23.2 
& 4.2  & 10.8 
& 7.3  & 19.7 
& 6.2  & 16.1 
& 5.1  & 14.5 
& 5.1  & 14.5 
& 6.7  & 21.4 \\
ATM-S \cite{li2024visual}
& 10.5 & 26.8 
& 7.1 & 24.8 
& 11.9 & 33.8 
& 14.7 & 39.4 
& 7.0 & 23.9 
& 11.1 & 35.8 
& 16.1 & 43.5 
& 15.0 & 40.3 
& 4.9 & 22.7 
& 20.5 & 46.5 
& 11.8 & 33.7 \\

UBP \cite{Wu2025UBP}
& 11.5 & 29.7 
& 15.5 & 40.0 
& 9.8 & 27.0 
& 13.0 & 32.3 
& 8.8 & 33.8 
& 11.7 & 31.0 
& 10.2 & 23.8 
& 12.2 & 32.2 
& 15.5 & 40.5 
& 16.0 & 43.5 
&12.4 & 33.4 \\

\textbf{SFTG(Ours)}
& 12.1 & 32.7 
& 17.2 & 29.4 
& 10.8 & 29.6 
& 14.4 & 36.5 
& 9.0 & 37.0 
& 12.4 & 35.6 
& 11.9 & 26.8 
& 14.2 & 35.9 
& 16.3 & 44.1 
& 17.9 & 45.9 
& \textbf{13.6} & \textbf{35.4} \\

\bottomrule
\end{tabular}
}
\label{tab:performance}
\end{table*}
\subsection{EEG Graph Transformer}
As our goal is to capture discriminative EEG features for brain decoding, it is essential to consider two unique properties of EEG signals: (1) Spatial brain connectivity, which can be inferred from the relationships between adjacent EEG electrodes; (2) Temporal neural dynamics, which are typically characterized by functional interactions among different brain regions over time \cite{hameed2024temporal}. 

From a graph perspective, we treat each EEG electrode as a node representing a specific brain region and propose to integrate both spatial and temporal relation learning into a full-relation learning framework for EEG graphs. This approach enables the aggregation of key neural connectivity and dynamic features from EEG representations. To achieve this, we introduce the EEG Graph Transformer (EGT), designed to effectively model spatial and temporal dependencies in EEG graphs (as illustrated in Fig. 2).

Given an EEG graph \( G_t(V_t, E_t) \) at time step \( t \), we leverage the predefined graph structure to generate positional encodings for EEG electrodes using the graph Laplacian:

\begin{equation}
\Delta = \mathbf{I} - \mathbf{D}^{-1/2} \mathbf{A} \mathbf{D}^{-1/2} = \mathbf{U}^T \mathbf{\Lambda} \mathbf{U},  
\label{eq_1}
\end{equation}

where \( \mathbf{A} \) and \( \mathbf{D} \) denote the adjacency and degree matrices of the EEG graph, respectively. The matrices \( \mathbf{\Lambda} \) and \( \mathbf{U} \) represent the eigenvalues and eigenvectors obtained from the Laplacian decomposition. Since EEG graphs within the same dataset often share an identical adjacency structure, we use \( \mathbf{A} \) to represent \( A_t \) for simplicity. Following \cite{dwivedi2020generalization}, we utilize the \( K \) smallest non-trivial eigenvectors as positional encodings for each node (electrode), denoted as \( \lambda_i \in \mathbb{R}^K \). These positional encodings are then transformed into a feature space of dimension \( d \) using an affine transformation:

\begin{equation}
\boldsymbol{h}_{i} = (\mathbf{W}_{v} \boldsymbol{v}_{i} + \boldsymbol{b}_{v}) + (\mathbf{W}_{p} \boldsymbol{\lambda}_{i} + \boldsymbol{b}_{p}),  
\label{eq_2}
\end{equation}

where \( \boldsymbol{h}_i \in \mathbb{R}^d \) is the position-enhanced node representation, \( \mathbf{W}_v \in \mathbb{R}^{d \times 3} \), \( \mathbf{W}_p \in \mathbb{R}^{d \times K} \), and \( \boldsymbol{b}_v, \boldsymbol{b}_p \in \mathbb{R}^d \) are learnable transformation parameters. The addition of positional encoding ensures that nearby electrodes exhibit similar representations, whereas distant electrodes are assigned more distinct features, thus improving EEG graph-based learning.

To capture functional relationships among EEG electrodes, we introduce multiple full-relation (FR) heads. These heads establish electrode interactions and update node representations by aggregating information from correlated nodes:

\begin{equation}
\boldsymbol{w}_{i, j}^{k,l} = \operatorname{Softmax}_j\left(\frac{(\boldsymbol{Q}^{k, l} \boldsymbol{h}_{i}^{(l)}) \cdot (\boldsymbol{K}^{k, l} \boldsymbol{h}_{j}^{(l)})}{\sqrt{d_k}}\right),  
\label{eq_3}
\end{equation}

\begin{equation}
\boldsymbol{\hat{h}}_i^{(l)} = \boldsymbol{O}^{l} {\bigg \|}_{k=1}^{H} \left(\sum_{j \in \mathcal{N}_i} \boldsymbol{w}_{i,j}^{k, l} \boldsymbol{V}^{k, l} \boldsymbol{h}_{j}^{(l)}\right),  
\label{eq_4}
\end{equation}

where \( \mathbf{Q}^{k,l}, \mathbf{K}^{k,l}, \mathbf{V}^{k,l} \in \mathbb{R}^{d_k \times d} \) are learnable matrices for query, key, and value transformations in the \( k \)-th FR head of the \( l \)-th layer. \( \mathbf{O}^l \in \mathbb{R}^{d \times d} \) is the output transformation matrix, \( \sqrt{d_k} \) is the scaling factor for normalized dot-product similarity, and \( w_{i,j}^{k,l} \) represents the interaction weight between electrodes \( i \) and \( j \) captured by the \( k \)-th FR head in the \( l \)-th layer. The operation \( {\bigg \|} \) denotes feature concatenation across different heads, with \( H \) representing the number of FR heads.

Our approach generalizes multi-head attention mechanisms \cite{vaswani2017attention} for EEG graph learning, allowing the model to effectively encode both spatial and functional relationships among electrodes. By employing multiple FR heads, the model can jointly attend to various connectivity patterns and extract highly informative features.

To refine the learned representations, we apply a Feed Forward Network (FFN) with residual connections \cite{he2016deep} and batch normalization \cite{ioffe2015batch}:

\begin{equation}
\boldsymbol{\overline{h}}_i^{(l)} = \operatorname{Norm}\left(\boldsymbol{h}_{i}^{(l)} + \boldsymbol{\hat{h}}_{i}^{(l)}\right),  
\label{eq_5}
\end{equation}

\begin{equation}
\boldsymbol{h}_i^{(l+1)} = \operatorname{Norm}\left(\boldsymbol{\overline{h}}_i^{(l)} + \mathbf{W}_{2}^{l} \sigma \left(\mathbf{W}_{1}^{l} \boldsymbol{\overline{h}}_i^{(l)}\right)\right).  
\label{eq_6}
\end{equation}

Here, \( \operatorname{Norm}(\cdot) \) represents batch normalization, \( \mathbf{W}_1^l \in \mathbb{R}^{2d \times d} \), \( \mathbf{W}_2^l \in \mathbb{R}^{d \times 2d} \) are learnable parameters, and \( \sigma(\cdot) \) is the ReLU activation function. The representations \( \boldsymbol{h}_i^{(l)} \) and \( \boldsymbol{h}_i^{(l+1)} \) denote the intermediate and final outputs for the \( l \)-th layer, respectively. Finally, to obtain a robust sequence-level EEG graph representation, we average the node features within each EEG graph and aggregate multiple consecutive graph representations:

\begin{equation}
\boldsymbol{S} = \frac{1}{f} \sum_{t=1}^{f} \boldsymbol{s}^{t} = \frac{1}{f} \sum_{t=1}^{f} \frac{1}{J} \sum_{i=1}^{J} \boldsymbol{h}^{t}_{i},  
\label{eq_7}
\end{equation}

where \( \boldsymbol{S} \) and \( \boldsymbol{s}^{t} \in \mathbb{R}^{d} \) represent the sequence-level and individual EEG graph representations respectively. For simplicity, we denote \( \boldsymbol{h}^{t}_{i} \) as the encoded representation of electrode \( i \) at time step \( t \). In this formulation, each electrode's representation, enriched with functional and spatial features (as shown in Eqs. \eqref{eq_3}, \eqref{eq_4}), is assumed to contribute equally to the overall graph representation, while each EEG graph is treated as equally important in capturing an individual's neural activity patterns.

\begin{figure*}
    \centering
    \includegraphics[width=0.95\linewidth]{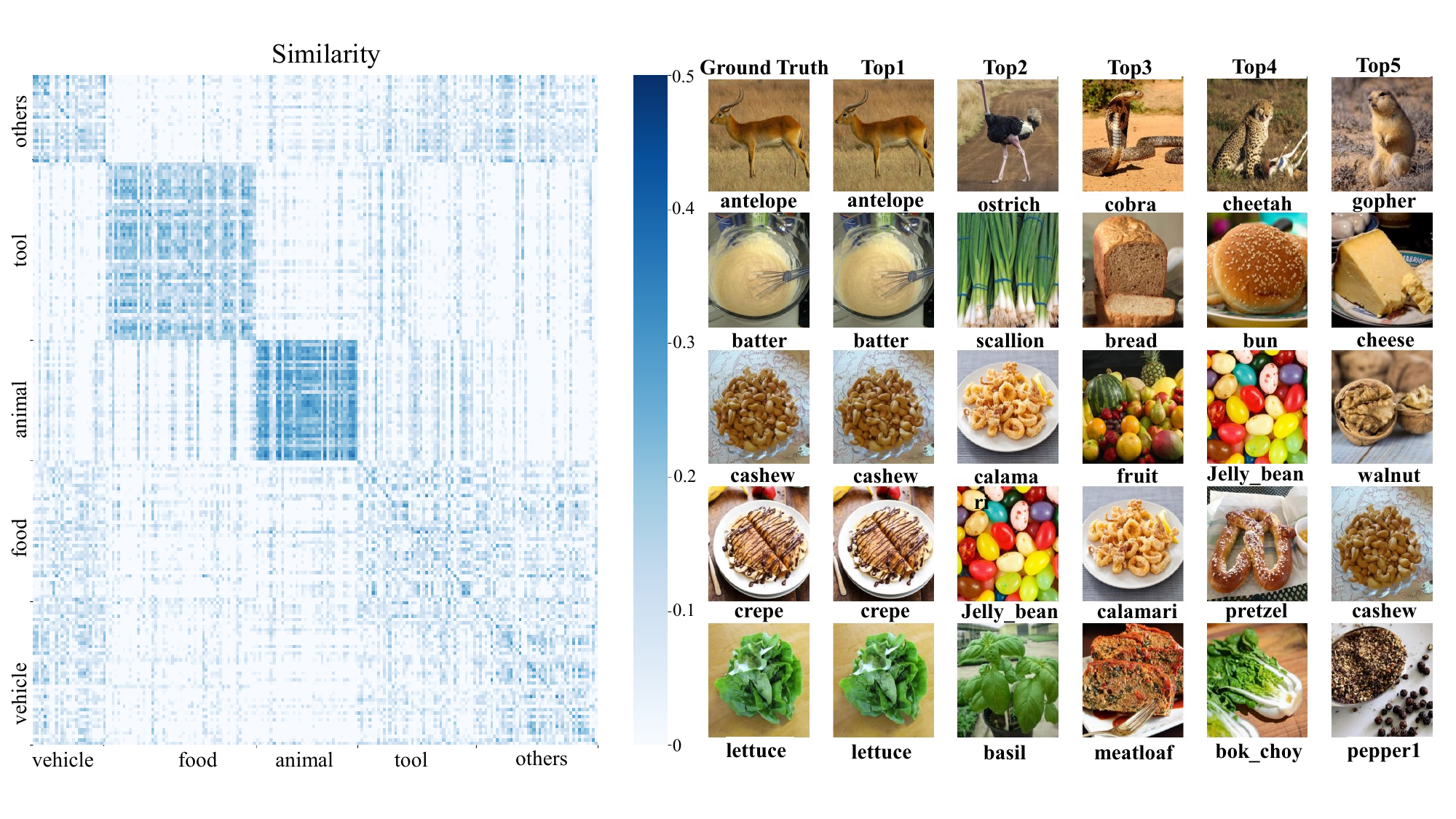}
    \caption{Semantic Similarity Analysis and Visualization. (A) shows the cosine similarity computed for feature pairs of 200 test-set concepts. (B) displays the classification outcomes, with the ground truth presented in the first column and the top-5 predictions in the subsequent columns.
}
    \label{fig:similarity}
\end{figure*}
\subsection{Graph archetype Contrastive Learning}
EEG signals are inherently prone to noise and artifacts during data acquisition, as highlighted in \cite{wagh2022evaluating}. This variability can result in multiple representations of the same underlying neural pattern. Furthermore, given the periodic nature of physiological signals, even minor temporal shifts may yield different manifestations of an otherwise consistent pattern \cite{haokunvector}. Consequently, it is imperative to consolidate these variations based on their shared generative origins to ensure stable learning signals for subsequent modeling.

Within this context, extracting the most representative EEG features (referred to as "archetype") for each individual becomes essential for capturing discriminative neural patterns. The GAC designs is inspired by previous approach involves clustering sequence-level representations to derive prototype, as demonstrated in \cite{wang2024graph}. However, this method typically produces identity-agnostic (i.e., pseudo-labeled) archetype with substantial uncertainty; significant intra-individual variability can lead to different sequences from the same individual being erroneously assigned to separate prototype groups. To address this limitation, we propose employing the archetype of the graph feature of each ground truth identity as the archetype of the EEG graph. These archetypes are then contrasted with the features of the sequence- and channel-level EEG graphs, thereby facilitating the learning of robust and discriminative representations for the identification of individuals based on EEG.

Given the encoded graph representations \( \{S^T_i\}_{i=1}^{N_1} \) derived from training EEG sequences \( \{X^T_i\}_{i=1}^{N_1} \), we categorize them based on ground-truth identities, forming the grouped set \( \{\mathbf{bS}_k\}_{k=1}^{C} \), where
\( \mathbf{bS}_k = \{ S_{k,j} \}_{j=1}^{n_k} \).

denotes the collection of graph representations belonging to the \( k \)-th subject, \( S_{k,j} \) is the sequence-level graph representation of the \( j \)-th EEG channel, and \( n_k \) is the total number of EEG sequences associated with the \( k \)-th identity. To derive a representative EEG graph archetype, we compute the archetype of the graph features within the same identity class:

\begin{equation}
\boldsymbol{c}_{k} = \frac{1}{n_k} \sum_{j=1}^{n_k} \boldsymbol{S}_{k,j},  
\label{eq_8}
\end{equation}

where \( \boldsymbol{c}_k \in \mathbb{R}^d \) represents the archetype feature for the \( k \)-th identity. To enhance the discriminative ability of EEG graph representations at both sequence and channel levels, we introduce the EEG Graph archetype Contrastive (GAC) Loss, defined as:

\begin{equation}
\mathcal{L}_{\mathrm{GAC}} = \alpha \mathcal{L}^{\text{seq}}_{\mathrm{GAC}} + (1 - \alpha) \mathcal{L}^{\text{ch}}_{\mathrm{GAC}},  
\label{eq_9}
\end{equation}

where:

\begin{equation}
\mathcal{L}^{\text{seq}}_{\mathrm{GAC}} = \frac{1}{N_1} \sum_{k=1}^{C} \sum_{j=1}^{n_k} -\log \frac{\exp \left (\boldsymbol{S}_{k, j} \cdot \boldsymbol{c}_{k} / \tau_1\right )}{\sum_{m=1}^{C} \exp \left (\boldsymbol{S}_{k, j} \cdot \boldsymbol{c}_{m} / \tau_1\right )},  
\label{eq_10}
\end{equation}

\begin{equation}
\resizebox{\columnwidth}{!}{$
\mathcal{L}^{\text{ch}}_{\mathrm{GAC}} = \frac{1}{fN_1} \sum_{k=1}^{C} \sum_{j=1}^{n_k} \sum_{t=1}^{f}  
- \log \Biggl( \frac{\exp\Bigl( \frac{\mathcal{F}_1\bigl(\boldsymbol{s}^{t}_{k,j}\bigr)
\cdot \mathcal{F}_2\bigl(\boldsymbol{c}_{k}\bigr)}{\tau_2} \Bigr)}
{\sum_{m=1}^{C} \exp\Bigl( \frac{\mathcal{F}_1\bigl(\boldsymbol{s}^{t}_{k,j}\bigr)
\cdot \mathcal{F}_2\bigl(\boldsymbol{c}_{m}\bigr)}{\tau_2} \Bigr)} \Biggr).
$}
\label{eq_11}
\end{equation}

In Eq.\eqref{eq_9}, \( \alpha \) is the balancing coefficient that fuses sequence-level (\( \mathcal{L}^{\text{seq}}_{\mathrm{GAC}} \)) and channel-level (\( \mathcal{L}^{\text{ch}}_{\mathrm{GAC}} \)) graph archetype contrastive learning. In Eqs.\eqref{eq_10} and \eqref{eq_11}, \( \boldsymbol{c}_m \) represents the archetype feature of the \( m \)-th identity, and \( \boldsymbol{s}^{t}_{k,j} \) denotes the graph representation of the \( t \)-th EEG channel corresponding to \( S_{k,j} \) of the \( k \)-th identity. The parameters \( \tau_1 \) and \( \tau_2 \) are temperature scaling factors for contrastive learning, while \( \mathcal{F}_1(\cdot) \) and \( \mathcal{F}_2(\cdot) \) are linear projection layers that align EEG channel representations and graph archetypes into the same contrastive space \( \mathbb{R}^d \).

It is important to note that EEG graph archetypes are computed from higher-level (i.e., sequence-level) representations, and the learnable linear projection in Eq.\eqref{eq_11} enables the model to integrate meaningful representations across both levels. The proposed GAC loss serves as a generalized EEG archetype contrastive learning approach, incorporating functional connectivity modeling along with channel-level and sequence-level graph learning (Eq.\eqref{eq_9}). This formulation can be theoretically interpreted as an Expectation-Maximization \cite{moon1996expectation} optimization strategy.

\section{Experiments}

\begin{figure*}
    \includegraphics[width=0.95\linewidth]{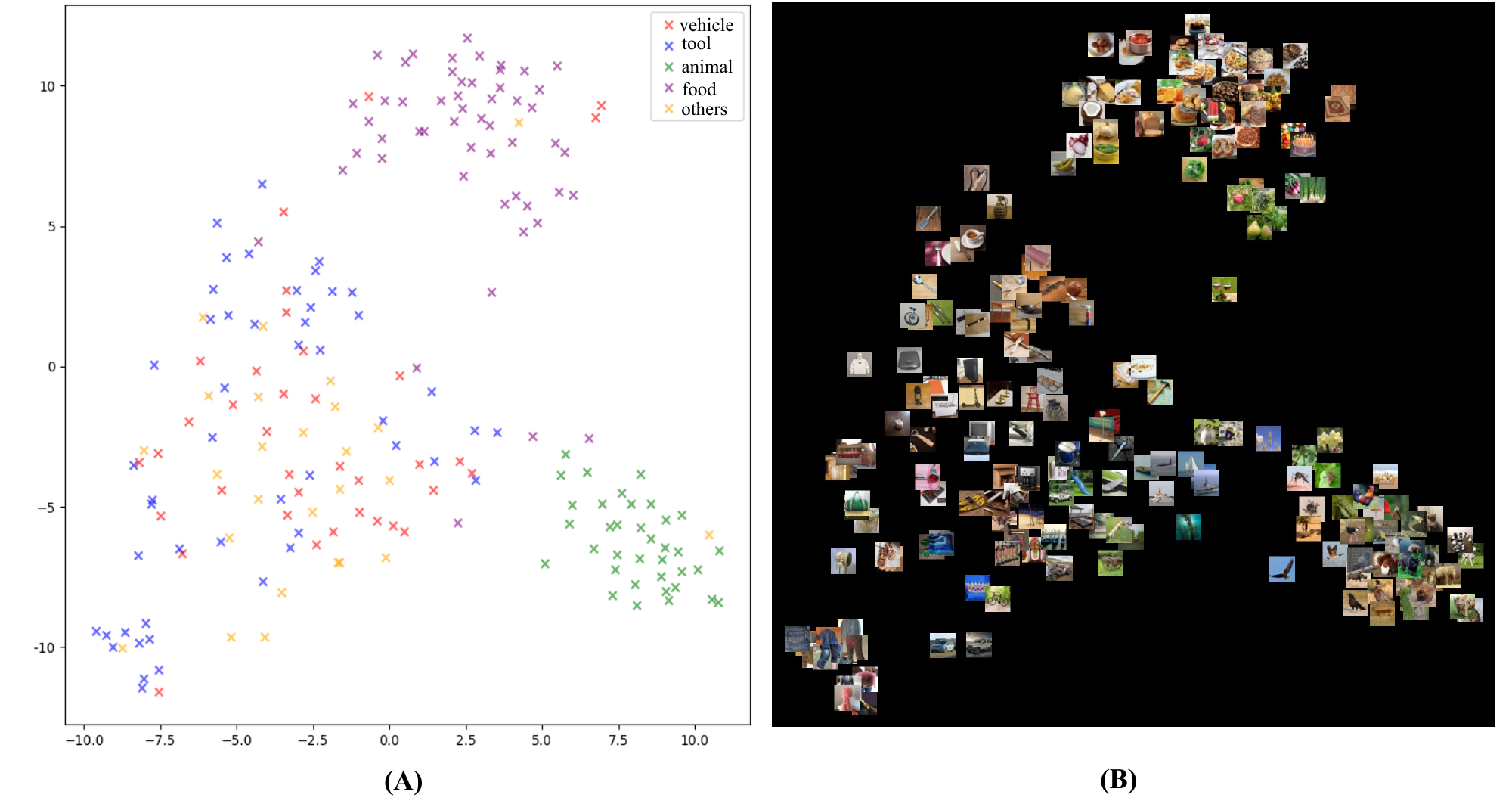}
    \caption{t-SNE visualization of five categories: animal, food, vehicle, tool, and others. (A) EEG feature distribution of SFTG training results. (B) Corresponding images in the t-SNE space.}
    \label{fig:tsne}
\end{figure*}

Table~\ref{tab:performance} reports the top-1 and top-5 accuracy for 200-way zero-shot classification under both \textit{subject-dependent} and \textit{subject-independent} settings. We compare our proposed model against several competitive baselines, including BraVL~\cite{du2023decoding}, NICE variants~\cite{song2024decoding}, ATM-S~\cite{li2024visual}, VE-SDN~\cite{chen2024visual}, and UBP \cite{Wu2025UBP}. More implementation details can be found in appendix.

In the subject-dependent setting, our method achieves the highest performance across nearly all subjects, with an average top-1 accuracy of \textbf{53.2\%} and top-5 accuracy of \textbf{82.4\%}, outperforming all the previous best method. In the subject-independent setting, where the model is trained on all but one subject and evaluated on the held-out subject, our model also achieves the best average performance: \textbf{13.7\%} top-1 and \textbf{38.8\%} top-5 accuracy.

\subsection{Qualitative analysis}

We conducted representational similarity analysis (RSA) following~\cite{cichy2020eeg}. We averaged the data across all subjects and computed similarity matrices for 200 test concepts grouped into five semantic categories: animal, food, vehicle, tool, and others. As shown in Fig.~\ref{fig:similarity}, EEG representations exhibit clear intra-category clustering, aligning closely with corresponding image features. We further employ t-SNE~\cite{van2008visualizing} to project EEG feature representations into a two-dimensional space, as visualized in Fig.~\ref{fig:tsne}. To enhance interpretability, we overlay the corresponding images at each embedding location. Our model yields clearly separated clusters, particularly for the animal and food categories.

\subsection{Ablation study}

To evaluate the effectiveness of each component in our proposed method. We consider a baseline model that processes raw EEG channel signals as independent features, without explicitly modeling cross-channel relationships. Then, we define naive contrastive learning (NC), which learns EEG representations without leveraging graph structures.As summarized in Table~\ref{tab:ablation}, introducing the EGT significantly outperforms both the baseline and AC models, demonstrating the importance of modeling EEG channels as a structured graph. 
Furthermore, incorporating GAC into EGT surpasses the EGT with naive contrastive learning (EGT + NC), which relies solely on cross-entropy loss. Notably, EGT combined with GAC outperforms the variant using direct contrastive learning (i.e., “EGT + DC” with cross-entropy loss) by achieving improvements of 1.1–3.4\% in mAP and 0.8–2.2\% in Rank-1 accuracy, verifying the crucial role of GAC in learning discriminative and representative graph features.

\begin{table}[t]
\centering
\caption{Ablation study with different configurations: Naive Contrastive Learning (NC), EEG Graph Transformer (EGT) with naive Contrastive Learning (NC) or Graph Archetype Contrastive Learning (GAC). “+” indicates that the component is employed.}
\label{tab:ablation}
\resizebox{\linewidth}{!}{%
\begin{tabular}{l|cc|cc|cc|cc|cc}
\toprule
\textbf{Configurations} & \multicolumn{2}{c|}{\textbf{Sub 1}} & \multicolumn{2}{c|}{\textbf{Sub 2}} & \multicolumn{2}{c|}{\textbf{Sub 3}} & \multicolumn{2}{c|}{\textbf{Sub 4}} & \multicolumn{2}{c}{\textbf{Sub 5}} \\
 & \textbf{mAP} & \textbf{R1} & \textbf{mAP} & \textbf{R1} & \textbf{mAP} & \textbf{R1} & \textbf{mAP} & \textbf{R1} & \textbf{mAP} & \textbf{R1} \\
\midrule
Baseline           & 5.8  & 5.3 & 4.9 & 4.2 & 5.5  & 4.7 & 6.4  & 5.8 & 5.4  & 4.9 \\
NC                 & 11.3 & 8.1 & 9.8 & 6.2 & 9.5 & 6.8 & 9.4 & 7.3 & 7.2 & 5.3 \\
EGT + NC           & 29.3 & 20.4 & 20.1 & 14.7 & 21.6 & 17.0 & 31.1 & 21.5 & 25.7 & 17.8 \\
EGT + GAC          & 32.7 & 22.6 & 21.2 & 15.5 & 22.3 & 17.5 & 33.1 & 22.8 & 27.0 & 18.6 \\
\bottomrule
\end{tabular}
}
\end{table}

\section{Conclusions}
In this paper, we propose \textbf{SFTG} to learn effective representations from EEG channel graphs for decoding visual neural representation. We design a transformer-based graph architecture to aggregate channels physical topology and functional connection into robust graph representations, capturing both local and global inter-dependencies. Experimental results show \textbf{SFTG} outperforms existing state-of-the-art models and can be effectively scaled to diverse subjects.

%Bibliography
\bibliographystyle{unsrt}  
\bibliography{reference}

\end{document}